\documentclass[letterpaper, 10 pt, conference]{ieeeconf}

\IEEEoverridecommandlockouts                              % This command is only needed if 
\usepackage{graphicx}
\usepackage{tabularx}
\usepackage{algorithm}
\usepackage{algpseudocode}
\usepackage{amsmath}

\newcolumntype{P}[1]{>{\centering\arraybackslash}p{#1}}
\newcolumntype{M}[1]{>{\centering\arraybackslash}m{#1}}

\newcommand{\norm}[1]{\left\lVert#1\right\rVert}

\usepackage{multirow}

\title{\LARGE \bf
Towards Automatic Annotation for Semantic Segmentation \\ in Drone Videos}

\author{Alina Marcu$^{1}$ Drago\c{s} Costea$^{2}$ Vlad Lic\u{a}re\c{t}$^{3}$ and Marius Leordeanu$^{4}$% <-this % stops a space
\thanks{$^{1}$ Alina Marcu is with University "Politehnica" of Bucharest and Mathematics Institute of the Romanian Academy
        {\tt\small alina.marcu@acs.stud.upb.ro}}%
\thanks{$^{2}$ Dragos Costea is with University "Politehnica" of Bucharest
        {\tt\small dragos.costea@acs.stud.upb.ro}}%
\thanks{$^{3}$ Vlad Licaret is with University "Politehnica" of Bucharest 
        {\tt\small vlad.licaret@etti.stud.upb.ro}}%
\thanks{$^{4}$ Marius Leordeanu is with University "Politehnica" of Bucharest and Mathematics Institute of the Romanian Academy
        {\tt\small marius.leordeanu@cs.pub.ro}}%
}

\begin{document}

\maketitle
\thispagestyle{empty}
\pagestyle{empty}

%%%%%%%%%%%%%%%%%%%%%%%%%%%%%%%%%%%%%%%%%%%%%%%%%%%%%%%%%%%%%%%%%%%%%%%%%%%%%%%%
\begin{abstract}

Semantic segmentation is a crucial task for robot navigation and safety. However, it requires huge amounts of pixelwise annotations to yield accurate results. While recent progress in computer vision algorithms has been heavily boosted by large ground-level datasets, the  labeling  time  has  hampered  progress  in  low  altitude  UAV applications, mostly due to the difficulty imposed by large object scales and pose variations. Motivated by the  lack of a large video aerial dataset, we introduce a new one, with high resolution (4K) images and manually-annotated dense labels every 50 frames. To  help  the video labeling  process, we make an important step towards automatic annotation and propose \textit{SegProp}, an  iterative  flow-based  method with geometric constrains to propagate the semantic labels to frames that lack human annotations. This results in a dataset with more than 50k annotated frames - the  largest  of its  kind,  to  the  best  of our  knowledge. Our experiments show that SegProp surpasses current  state-of-the-art label propagation methods  by  a  significant  margin. Furthermore, when training a semantic segmentation deep neural net using the automatically annotated frames,
we obtain a compelling overall performance boost at test time of 16.8\% mean F-measure over a baseline trained only with manually-labeled frames. 
The dataset, the label propagation  code and  a  fast  segmentation  tool  will  be made publicly  available.
\end{abstract}

%%%%%%%%%%%%%%%%%%%%%%%%%%%%%%%%%%%%%%%%%%%%%%%%%%%%%%%%%%%%%%%%%%%%%%%%%%%%%%%%
\section{INTRODUCTION}

The ability to anticipate events in the near future is a critical attribute for real-time autonomous systems and should be based on understanding the world scene at the semantic level. Visual semantic segmentation, which addresses the problem of localizing and identifying the different object categories in a given scene, is a precursor to any kind of action involving such objects, from localizing and moving towards them to various, possibly complex, interactions. Even without the help of depth or other information (such as optical flow), people have very good accuracy in segmenting images into visual categories. Such task remains a challenge for robots.

While ground vehicles are forced to move bidirectionally, aerial robots are free to navigate in three dimensions. This allows them to capture images of objects from a wide range of scales and angles, with richer views than the ones available in datasets collected on the ground. Unfortunately, this unconstrained movement imposes significant challenges for accurate semantic segmentation, mostly due to the aforementioned variation in object scale and viewpoint. 

Classic semantic segmentation approaches focused on ground indoor and outdoor scenes. More recent work tackled imagery from the limited viewpoints of specialized scenes, such as ground-views of urban environments (from vehicles) and direct overhead views (from orbital satellites). Nevertheless, recent advances in aerial robotics allows us to capture previously unexplored viewpoints and diverse environments more easily. Given the current state of technology, in order to evaluate the performance of autonomous systems, the human component is considered a reference. However, human annotations are very expensive and especially in the context of videos, which have a huge number of frames, the ability to perform automatic annotation would be extremely valuable.

In this paper we introduce Ruralscapes, the largest high resolution (4K) video dataset for aerial semantic segmentation, taken in flight over rural areas in Eastern Europe. Then we start from a relatively small subset of humanly labeled frames in a video and perform SegProp, our novel iterative label propagation algorithm, to automatically annotate the whole sequence. Given a start and an end frame of a video sequence, SegProp finds pixelwise correspondences between labeled and unlabeled frames, to assign a class for each pixel in the video based on an iterative class voting procedure. In this way we generate huge amounts of labeled data (over 50k segmented frames) to use in training deep neural networks and show that the automatically labeled training frames help significantly in boosting the performance at test time.

Our pipeline can be divided into three steps. The first and most important is the data labeling step. We leverage the advantages of high quality 4K aerial videos, such as small frame-to-frame changes (50 frames per second) and manually annotate a relatively small fraction of frames, sampled at 1 frame per second. 
Then, we automatically generate a label for each intermediate frame between two labeled ones, using the SegProp algorithm (Sec. \ref{sec:algorithm}).
As final step, we mix the manually and automatically annotated frames and use them for training. 

%Our main objective is to develop a reliable solution for scene semantic segmentation, on a real-world scenario (video), with an UAV flying at various altitudes, with both static and dynamic objects, along with complex landscape in the background. The proposed solution can run in real-time on an embedded platform and is able to accurately outline objects from 12 different classes (Fig. \ref{fig:sample_label_detail}).

\begin{figure}
\centering
\includegraphics[scale=0.22,keepaspectratio]{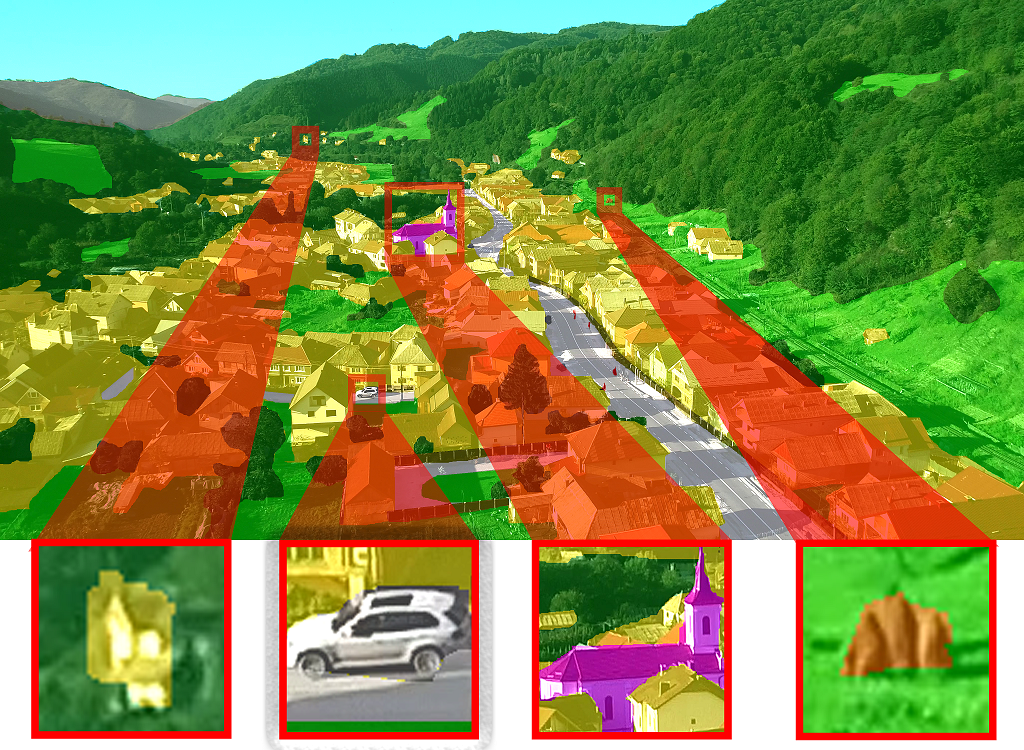}
\caption{\label{fig:sample_label_detail} Sample label image overlaid on top of its corresponding RGB image with detail magnification. Small classes such as haystack and car are difficult to segment accurately, but overall the labeled frames contain a very good level of detail. The dataset offers a large variation in object scale: classes generally easy to segment up close such as buildings turn into difficult classes far away from the camera.}
\vspace{-6mm}
\end{figure} 

\textbf{Datasets for semantic segmentation in video}. Since most work is focused on ground navigation, the largest datasets with real-world scenarios are ground-based. Earlier image-based segmentation datasets, such as Microsoft's COCO~\cite{lin2014microsoft}, contained rough labels, but the large number of images (123k) and classes (80), made it a very popular choice. 
Cityscapes~\cite{Cordts2016Cityscapes} was among the first large-scale dataset for ground-level semantic and instance segmentation. Year after year, the datasets increased in volume and task complexity, culminating with Apolloscape~\cite{huang2018apolloscape}, which is, to the best of our knowledge, the largest real ground-level dataset. Compared to its predecessors, it also includes longer video shots, not just snippets. It comprises of 74,555 annotated video frames. To help reduce the labeling effort, a depth and flow-based annotation tool is used.
Aeroscapes~\cite{nigam2018ensemble} is a UAV dataset that contains real-world videos and semantic annotations for each frame and it is closer to what we aim to achieve. Unfortunately, the size of the dataset is rather small, with video snippets ranging from 2 to 125 frames. It includes 3,269 sparsely labeled frames.
The most similar dataset to ours is UAVid ~\cite{lyu2018uavid}. It has about 10 times less pixels and despite being introduced a year ago, it is not yet public.

Since labeling real-world data (especially video) is difficult, a common practice is to use synthetic videos from a simulated environment. Such examples are Playing Playing for Benchmarks~\cite{richter2017playing}, for ground-level navigation and the recently released Mid-air~\cite{Fonder2019MidAir}, for low-altitude navigation. Mid-air has more than 420k training video frames. The diversity of the flight scenarios and classes is reduced - mostly mountain areas with roads - but the availability of multiple seasons and weather conditions is a plus.

%\subsection{Label propagation methods}

\textbf{Label propagation methods}. Recent methods for automatic label propagation need a single human annotated frame. That is, given one frame, they extend the label to nearby frames. The state-of-the-art results on Cityscapes and KITTI of SDCNet~\cite{zhu2019improving} confirms the advantage of the approach. Other authors try to use semi-supervised learning to improve the intermediate labels~\cite{budvytis2010label}.

The most similar method to our approach (propagate labels between two frames) is~\cite{budvytis2017large}, for ground navigation with low resolution images (320x240).
They employ an occlusion-aware algorithm coupled with an uncertainty estimation method, related to the label relaxation technique from~\cite{zhu2019improving}. Their code is not made public for direct comparison. Also their approach is less useful in our case, where we have very high resolution images at a high frame rate (50fps) and dense optical flow can be accurately computed.

\begin{figure}
\centering
\includegraphics[scale=0.33,keepaspectratio]{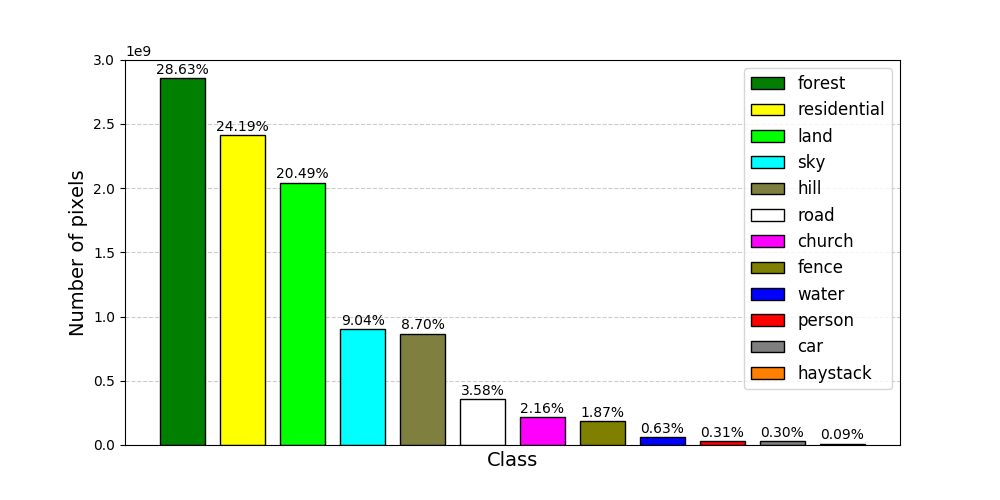}
\caption{\label{fig:pixel_distribution} Class pixels' distribution. Being a rural landscape, the dominant classes are buildings, land and forest (73.01\% combined). Due to the flight altitude, smaller classes such as haystack, car and person hold a very small percentage. Nevertheless, this distribution helps common UAV tasks such as mapping, navigation with obstacle avoidance and safe landing or more complex applications such as package delivery.}
\vspace{-6mm}
\end{figure}

\vspace{2mm}
In this paper we make the following \textbf{main contributions}:
\begin{itemize} 
\item We introduce Ruralscapes the largest high resolution (4K) video dataset for aerial semantic segmentation composed of 50,835 fully annotated frames with 12 semantic classes.
\item We propose an iterative, optical flow based label propagation method, termed SegProp, with geometric constraints, that outperforms similar state-of-the-art algorithms. 
\item We show that our method can easily integrate other similar label propagation methods in order to further improve the segmentation results.
\end{itemize}

\section{Ruralscapes: A Dataset for Rural UAV Scene Understanding with Large Altitude Changes}

\subsection{Manual annotation tool}

We designed a user-friendly tool that facilitates drawing the contour of objects (in the form of polygons). For each selected polygon we can assign one of the 12 available classes. The class set includes background objects such as forest, land, hill, sky, residential, road or river, and also, some foreground, countable objects, like person, church, haystack, fence and car.

We developed this tool mostly to speed up segmentation. Our software is suited for high resolution images. Furthermore, it offers support for hybrid contour/point segmentation - the user can alternate between point-based and contour-based segmentation during a single polygon. The most time-saving feature, assuming the image needs to be fully segmented (e.g., no 'other' class), is a 'send to back' functionality to copy the border from the already segmented class to the new one being drawn. Finally, it includes intuitive polygon editing capabilities (overlapping polygons are easy to select and modify). None of the existing tested solutions provided all of the above functionality~\cite{cocoannotator, russell2008labelme, dataturks, supervisely}. The software is portable (Python) and will be released alongside the dataset. %A screenshot is shown in Figure ~\ref{fig:tool_screenshot}.
\vspace{-1mm}
\subsection{Dataset details}

We have collected 20 high quality 4K videos portraying rural areas. Ruralscapes comprises of various landscapes, different flying scenarios at multiple altitudes and objects across a wide span of scales. The video sequence length varies from 11 seconds up to 2 minutes and 45 seconds. The dataset consists of 17 minutes of drone flight, resulting in 50,835 fully annotated frames with 12 classes. Of those, 1,047 were manually annotated. To the best of our knowledge, it is the largest dataset for semantic segmentation from real UAV videos.

Labels have a good level of detail. However, due to the small spatial resolution of the far away or small classes, accurate segmentation is difficult, as seen in the sample label from Figure \ref{fig:sample_label_detail}. Some classes, such as haystack, are very small by the nature of the dataset, others such as person, also feature close-ups. Based on the feedback received from the 21 people that segmented the dataset, it took them on average 45 minutes to label a single frame. This translates into 846 human hours needed to segment the manually labeled 1047 frames.

The distribution of classes in terms of occupied area is shown in Figure \ref{fig:pixel_distribution}. Background classes such as forest, land and residential are dominant, while smaller ones such as person and haystack are at the opposite spectrum.
Based on the feedback received from the people that helped with the labeling, small objects were the most difficult to segment. 

\begin{figure}
\centering
\includegraphics[scale=0.12,keepaspectratio]{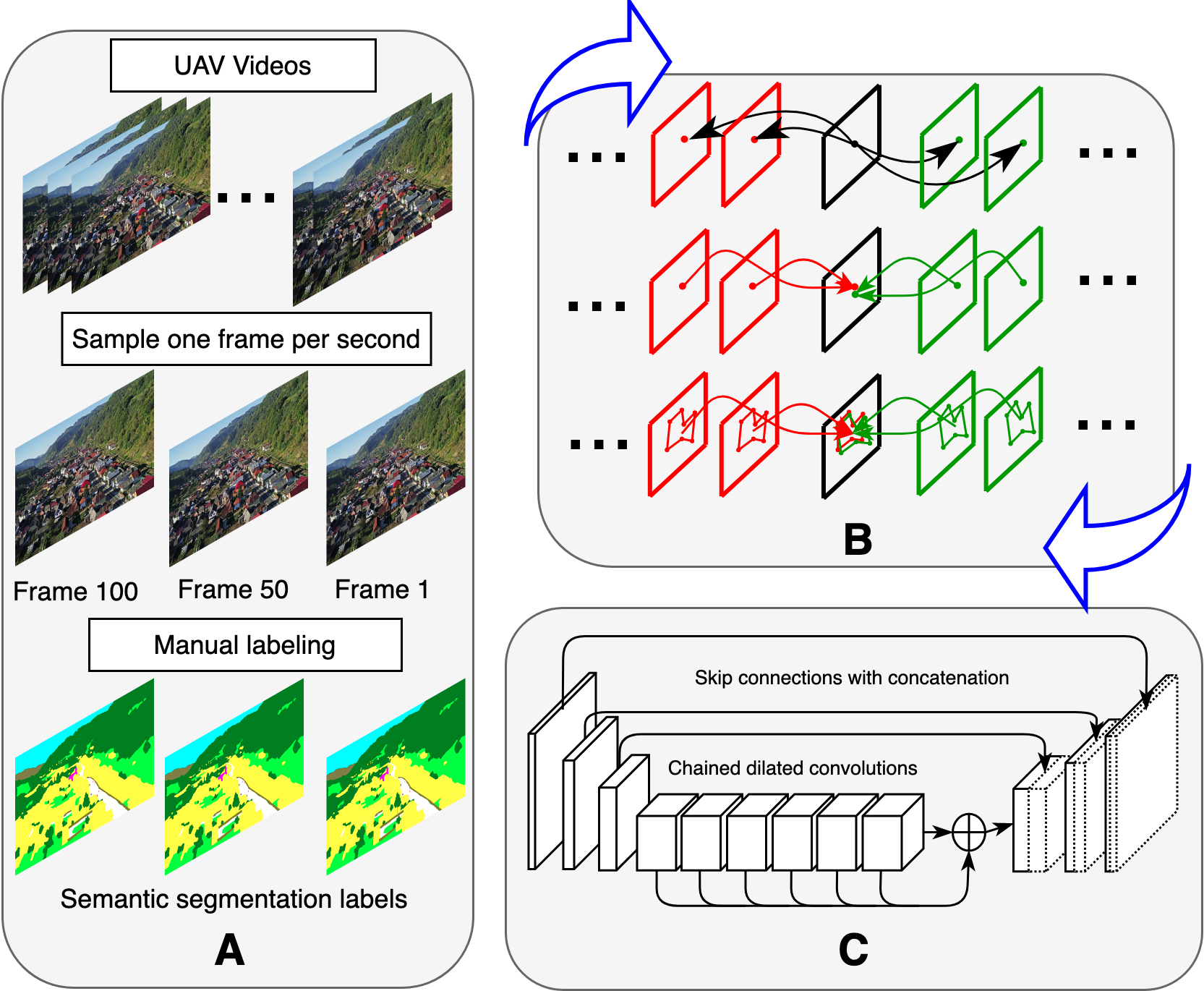}
\caption{\label{fig:overview} Overview of the proposed method for automatic propagation of semantic labels in the context of aerial semantic segmentation.
\textbf{A.} The UAV videos are sampled at one frame per second and the resulting frames are manually labeled. \textbf{B.} The labels were propagated to the remaining frames using our SegProp algorithm, based on class voting at the pixel level according to (1) forward and backward flow from the current frame to a manually annotated frame (2) region-based homography maps computed between current and manually labeled frames and (3) iterations of 1 and 2 among neighboring frames. \textbf{C.} All frames were used to train a UNet-like CNN with dilated convolutions~\cite{marcu2018safeuav}.}
\vspace{-6mm}
\end{figure} 

\section{AUTOMATIC LABEL PROPAGATION}
\label{sec:algorithm}

\subsection{SegProp: Automatic Label Propagation Algorithm}

We propose a flow-based label propagation method, summarized in Algorithm \ref{alg:labelInterpHomography} and discussed at a theoretical level in Sec. \ref{sec:math_interp}. Let $P_{k}$ be an intermediate video frame between two
manually-labeled frames $P_{i}$ and $P_{j}$. We extract optical flow both forward and backward ($F_{i\rightarrow j}$ and  $F_{j\rightarrow i}$) using PWCFlow~\cite{sun2018pwc} from RGB images. We then use the pixel motion trajectories from optical flow in order to map pixels from the annotated frames $P_{i}$ and $P_{j}$ to $P_{k}$ and vice-versa. This results in 4 correspondence maps,
two from $P_{k}$ to its manually labeled frames ($P_{k\rightarrow i}$,$P_{k\rightarrow j}$) and two from the labeled frames to $P_{k}$ ($P_{i\rightarrow k}$, $P_{j\rightarrow k}$), which can be used to place class votes from the manually labeled frames to the current unlabeled one.

\begin{figure}
\centering
\includegraphics[scale=0.22,keepaspectratio]{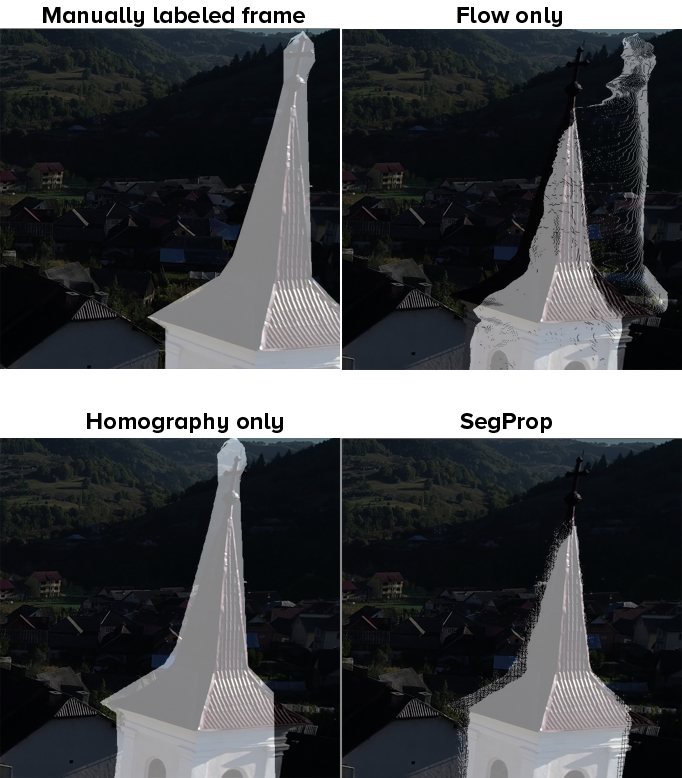}
\caption{\label{fig:interpolation_results} Label propagation results: RGB frame with manual white label overlaid, flow-based voting only, homography-based voting only, and full flow and homography combined voting propagation. While the homography based voting produces "cleaner" semantic regions, an agreement between optical flow and homography is desirable.}
\vspace{-7mm}
\end{figure}

Unfortunately, even state-of-the-art optical flow is prone to noise. In order to obtain a more robust voting, we incorporate geometric constrains - two additional votes are computed from regions transformed using homography estimation between regions in the the left and right labeled frames.
The homography based voting is particularly useful in edge preservation where the CNN-based optical flow generally lacks precision (Fig. \ref{fig:interpolation_results}). 
The labels in the ground truth segmentations
are first grouped into connected regions for each class. 
Then, a homography is computed using RANSAC for each connected component region from one bounding frame to the other one and viceversa ($P_{i\rightarrow j}$ and  $P_{j\rightarrow i}$). Labels are projected from the source frame to the destination and viceversa, while sending labeling votes to all intermediate frames in the process. In order for the transformation to yield accurate results, the region should "behave" like a planar one, which is especially true for distant regions.
We empirically find this approximation to yield more accurate votes than optical flow. Finally, the most voted class becomes the label of each pixel in $P_{k}$.

Even with the six votes, a certain degree of noise is still present. In order to further improve the labels, we propose the final iterative propagation method, SegProp, summarized in Algorithm ~\ref{alg:labelInterpIterations}. The main idea is that after the initial voting, to establish a more coherent agreement among neighboring frames by iteratively propagating class votes between each other using the same propagation procedure (see Algorithm ~\ref{alg:labelInterpHomography})
This approach results in better local consensus, generally translated in smoother and more accurate labels. 

\begin{algorithm}
\caption{Automatic label propagation with geometric constraints}
\label{alg:labelInterpHomography}
\begin{algorithmic}
    \State 1) Given labeled frames $P_i$ and $P_j$, consider an intermediate frame $P_k$.
    \State 2) Compute optical flow $F_{i\rightarrow j}$ and  $F_{j\rightarrow i}$ from RGB data.
    \State 3) Extract 4 class maps for the current frame $P_k$ by following pixel movements (according to optical flow) through time, where each pixel receives a corresponding class from the ground truth labels: $p_{k\overline{1,4}}(x,y) \leftarrow Class_{flow}$
      \State \hspace{1cm}2 forward ($P_{k\rightarrow i}$ , $P_{k\rightarrow j}$)
      \State \hspace{1cm}2 backward ($P_{i\rightarrow k}$ , $P_{j\rightarrow k}$),
    \State 4) Generate 2 additional class maps by computing homography transformations between connected components $CC$ (connected regions with the same class label) from $P_i$, $P_j$ and their flow correspondence:
      \For{each $CC$ in each class $C$}
            \For{$p_{CC}(x,y)$ in $P_k$}
            \State $p_{i \rightarrow k CC_{Cl}}(x,y) = p_{i flow}(x,y)$
            \State $p_{j \rightarrow k CC_{Cl}}(x,y) = p_{j flow}(x,y)$
            \EndFor
\vspace*{-4mm}
    \State $$H_{i \rightarrow j} \leftarrow RANSAC(p_{i CC_{Cl}}(x,y),p_{j CC_{Cl}}(x,y))$$
    \State $p_{k5 CC_{Cl}}(x,y) = H_{i \rightarrow j}(p_{i \rightarrow k CC_{Cl}}(x,y))$
    \vspace*{-2mm}

    \State $$H_{j \rightarrow i} \leftarrow RANSAC(p_{j CC_{Cl}}(x,y),p_{i CC_{Cl}}(x,y))$$
    \State $p_{k6 CC_{Cl}}(x,y) = H_{j \rightarrow i}(p_{j \rightarrow k CC_{Cl}}(x,y))$
    \EndFor
    \State 5) $class_k(x,y) = max(p_{k\overline{1,6}}(x,y))$
\end{algorithmic}
\end{algorithm}
%\vspace{-5mm}
\begin{algorithm}
\caption{SegProp Algorithm for Iterative Label Propagation}
\label{alg:labelInterpIterations}
\begin{algorithmic}
    \State 1) For a given frame $k$, perform steps \textbf{1-4} from \textbf{Algorithm 1} considering its neighboring $2f+1$ frames at distances 
     $i \in (1,f)$ and accumulate votes.
    \State 2) For each pixel vote for the majority class $class_k(x,y) = max(p_{k\overline{1,6\cdot f}}(x,y))$. Then go back to 1, until maximum number of iterations is reached.
\end{algorithmic}
\end{algorithm}

%In the case of the region-based homography-based voting (described in Algorithm \ref{alg:labelInterpHomography}), the intuition was that we could approximate regions of the same class to a plane and a region is visible in the one second that passes between two manually annotated frames. Given the optical flow correspondences between the current frame and a manually labeled anchor frame, 
%for pixels in a given region of the same class, we can estimate a homographic transformation which could bring two additional class votes, one for each forward/backward direction. We apply the homography and optical flow based voting scheme iteratively (Algorithm \ref{alg:labelInterpIterations}). The iterative scheme reduces the inter-frame class fluctuations and noises and generallyy increases the accuracy of labels.
%\vspace{-3mm}
\subsection{\label{sec:math_interp}Mathematical interpretation of our algorithm}

SegProp can be expressed mathematically as maximizing a certain clustering score:

\begin{equation} 
S_L = \mathbf{M}_{ia,jb} \cdot \mathbf{x}_{ia} \cdot \mathbf{x}_{jb},
\end{equation}

where $\mathbf{x}$ is an indicator vector that captures the segmentation such as:
\begin{equation*}
x_{ia} = \begin{cases} 
         1, & \text{if node $i$ has label class $a$} \\
         0, & \text{otherwise}
        \end{cases}.
\end{equation*}

and $\textbf{M}_{ia,jb}$ is the pairwise consistency between node $i$ and label $a$ and node $j$ and its label $b$. We can consider every pixel in the video as a node in a graph. For any node $i$, we can assign a label $a$, thus we have an unique index $ia$. Our mathematical interpretation is conceptual, in theory, as we never explicitly build $\textbf{x}$ or $\textbf{M}$. 

\begin{figure}
\centering
\includegraphics[scale=0.3,keepaspectratio]{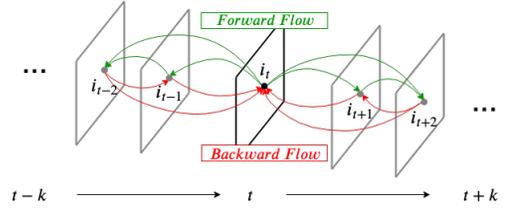}
\caption{\label{fig:graph_propagation} We could formulate SegProp as maximizing a clustering score over a graph in space and time. We consider a pixel at a given frame $t$ in the video as a node $i_{t}$ in our graph. Nodes are linked by optical flow (forward or backward) along a path of consecutive frames. Maximizing the clustering score produces labels that are consistent along flow paths in time.}
\vspace{-6mm}
\end{figure}

In the voting case, we consider only links between pairs of nodes (pixels at different time frames) that are put into correspondence by optical flow chains (by following the optical flow pixel movements from one frame to another), the estimated homography of whole class regions or any other mapping procedure (see Figure \ref{fig:graph_propagation} for a visual representation).
Thus $i \in N_{j}$  if and only if $i$ and $j$ are connected by such procedures. This way, we encourage connected pixels to have the same label by defining $\textbf{M}_{ia,jb}$:
\begin{equation*}
\mathbf{M}_{ia,jb} = \begin{cases} 
         1, & \text{if and only if a = b and $i \in N_{j}$ or $j \in N_{i}$} \\
         0, & \text{otherwise}
        \end{cases}.
\end{equation*}

Then SegProp can be mathematically defined as:
\begin{equation}
\mathbf{x}^{*} = \operatorname*{argmax}_x \sum_{ia} \sum_{jb} \mathbf{M}_{ia,jb} \cdot \mathbf{x}_{ia} \cdot \mathbf{x}_{jb}, 
\end{equation} where $\mathbf{x}_{ia} = 1$ if node $i$ has label $a$ and 0, otherwise.

In other words:
\begin{equation} \label{eq:3} S_{L}(x) = \mathbf{x}^{T} \cdot \mathbf{M} \cdot \mathbf{x}, \end{equation} and 
\begin{equation} \mathbf{x}^{*} = \operatorname*{argmax}(\mathbf{x}^{T} \cdot \mathbf{M} \cdot \mathbf{x}), \end{equation} with conditions $\sum_{a}(\mathbf{x}_{ia}) = 1$ and $\mathbf{x}_{ia} = \{0, 1\}$.

The relation to voting is immediate:
\begin{equation}\begin{aligned}
S_{L} & = \mathbf{x}^{T} \cdot \mathbf{M} \cdot \mathbf{x} \\
      & = \sum_{ia}\sum_{jb} \mathbf{M}_{ia,jb} \cdot \mathbf{x}_{ia} \cdot \mathbf{x}_{jb} \\
      & = \sum_{\substack{ia \\ j \in N_{i}}} \mathbf{M}_{ia, ja} \cdot \mathbf{x}_{ia} \cdot \mathbf{x}_{ja} \\
      & = \sum_{i}N_{i}(a),
\end{aligned}
\end{equation} where $N_{i}(a)$ are the number of neighbors of node $i$ that have the same label $a$ as not $i$. If $i$ is a node in a ground truth frame, $i$ has fixed label $a^*$. Maximizing the clustering score has a natural and intuitive meaning. We want to find the segmentation $x$ that encourages nodes with connections to have the same label. In the light of this mathematical formulation, one can show immediately that our iterative voting algorithm reduces to:

\begin{equation}
\mathbf{x}^{(t + 1)} = P_{L}(\mathbf{M} \cdot \mathbf{x}^{(t)}), \end{equation} where $P_{L}$ is a projection on the space of valid, feasible solutions. 

This result is directly related to classical inference methods in Markov Random Fields (MRFs)~\cite{li1994markov}. It can be interpreted as an instance of parallel Iterative Conditional Modes (ICM)~\cite{besag1986statistical}. That method is guaranteed to find a local optimum if done sequentially. However, if done in parallel it is faster and works well in practice.

Our algorithm is also related to the IPFP algorithm~\cite{leordeanu2009integer}, with the only difference being that we do not perform the optimal line search between $\mathbf{x}^{(t)}$ and $\mathbf{x}^{(t + 1)} = P_{L}(\textbf{M} \cdot \mathbf{x}^{(t)})$. This would be more difficult in our case, as we never explicitly work with $\mathbf{x}$ and $\mathbf{M}$ - the graph is only considered at a conceptual level. Computation and memory constraints would make it impossible to build $\mathbf{M}$ and $\mathbf{x}$ in practice, in order to optimize over the pure algebraic formulation. It is interesting to note that the projection $P_{L}$, which takes soft-valued segmentations $\mathbf{x}$ (i.e. votes) into the feasible domain of discrete labels, if replaced by a projection on a sphere $\norm{\mathbf{x}}_2 = 1$, it would transform SegProp into the classic Power Iteration for finding the main eigenvector of $\mathbf{M}$~\cite{mises1929praktische}. That formulation is known to solve the spectral clustering problem (one of its variants). 

The conceptual, mathematical interpretation of our algorithm is interesting. We believe that such formal equations can help in better understanding the properties of our algorithm and improving it both from theoretical and practical points of view.

\subsection{Training with automatically generated labels}

We trained an embeddable-hardware compatible system based on deep convolutional networks, specially designed for dense pixelwise prediction which has previously shown to yield good results on depth and safe landing area estimation using only the RGB input~\cite{marcu2018safeuav}. Our approach, however, is general and could work with any semantic segmentation method. The neural net model we use, termed SafeUAV-Net-Large, comprises of three down-sampling blocks followed by a chain of concatenated dilated convolutions, with progressively increasing dilation rates (1, 2, 4, 8, 16 and 32). Each dilated convolution outputs a set of 256 activation maps. The model is fully-convolutional and outputs a map with the same dimension as its input. This is done with three up-scaling blocks.
Each down-sampling block has two convolutional layers with stride 1, followed by a $2\times2$ max-pooling layer. Each up-scaling layer has a transposed convolution layer, a feature map concatenation with the corresponding map from the down-sampling layers and two convolutional layers with stride 1. The number of feature maps double after each down-sampling block, starting from 32 and halve for the up-sampling ones. Each convolution in the model has kernels of size $3\times3$. A visual representation of the architecture is portrayed in Figure \ref{fig:overview} \textbf{C}.

\begin{figure*}[t]
\begin{center}
\includegraphics[scale=0.26]{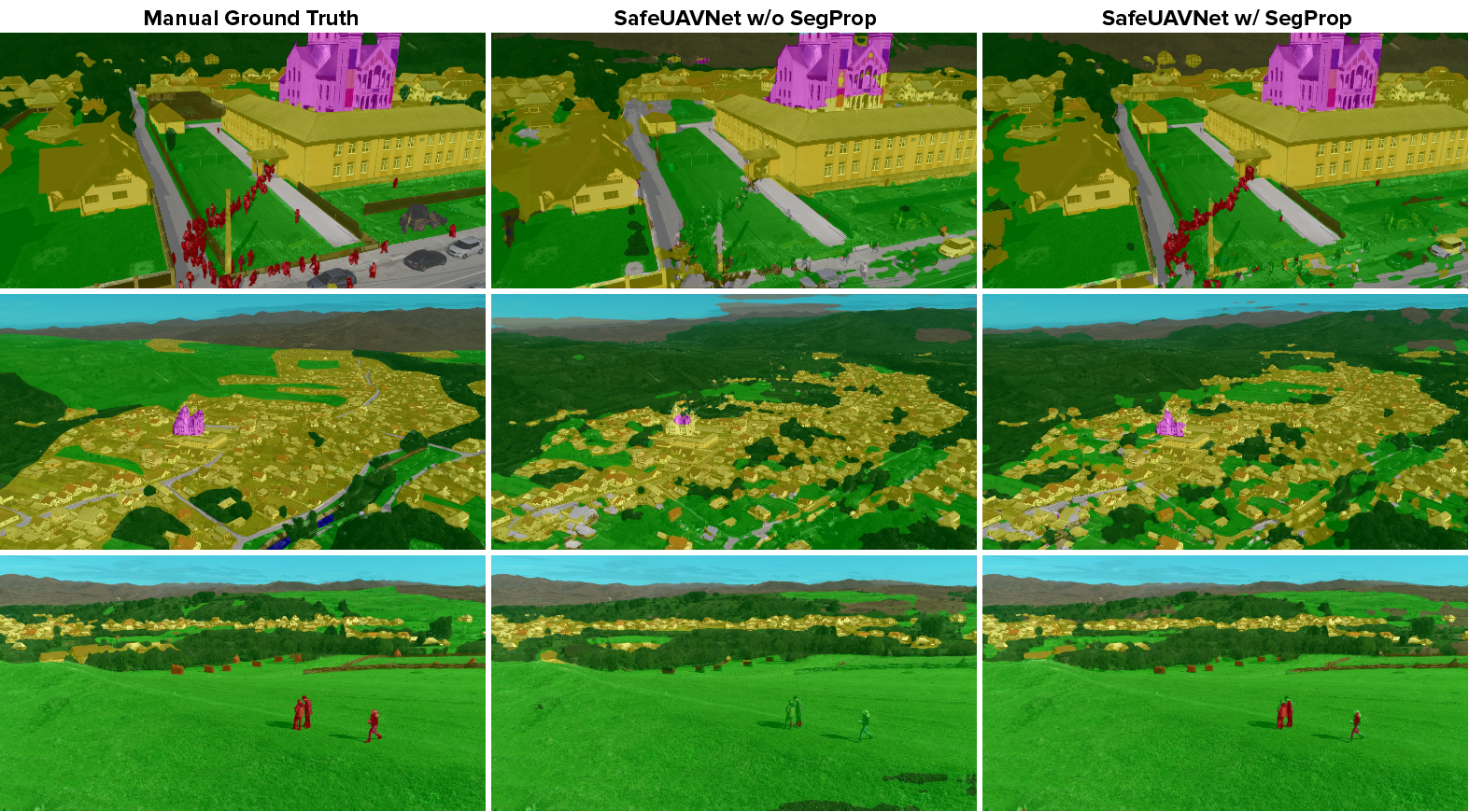}
   \caption{\label{fig:qualitative_results}Qualitative results on the testing set. SegProp helps both small classes (person, haystack) as well as large classes (an example above is the sky and forest from the second row and the land in the background of the third row. Thus, not only the small classes are  better represented, but the large ones also benefit from a more spatially coherent detection - e.g., the grass close to the humans in the third row.}
\end{center}
\vspace*{-3mm}
\end{figure*}

\section{EXPERIMENTAL ANALYSIS}
\begin{table*}[t]
%\vspace{-5mm}
\caption{\label{table:experimental_results}Neural network training results. SegProp provides a significant performance boost over the baseline. We report mean F-measure over all videos from the testing set, for each individual class.}
%The (*) results denote additional post processing of the prediction with SegProp. While the numbers are comparable, it results in better temporal consistency of the prediction maps.
\vspace{-5mm}
\label{table_results}
\begin{center}
\begin{tabular}{|c|c|c|c|c|c|c|c|c|c|c|c|c|c|}
\hline
Methods & Land & Forest & Residential & Haystack & Road & Church & Car & Water & Sky & Hill & Person & Fence & Overall\\
\hline
w/ Base Train & .495 & .496 & .774 & .000 & .252 & .166 & .000 & .006 & .952 & .371 & .000 & .060 & .298\\
w/ SegProp Train & \textbf{.540} & \textbf{.516} & \textbf{.822} & .586 & \textbf{.432} & \textbf{.382} & \textbf{.066} & .146 & \textbf{.985} & \textbf{.407} & \textbf{.471} & \textbf{.233} & \textbf{.466}\\
%w/ SegProp Train* & .539 & .514 & .818 & \textbf{.588} & \textbf{.432} & .377 & .062 & \textbf{.147} & .984 & .402 & .470 & \textbf{.233} & .464\\
\hline
\end{tabular}
\end{center}
\vspace{-5mm}
\end{table*}

\subsection{Dataset split}
The whole 20 densely labelled video sequences are divided intro training and testing video subsets. We used 7 different testing videos ($\approx$29.61\% of the total frames from the dataset) for evaluating the performance of our methods. The testing set consists of 311 manually-labeled frames and a total of 15,051 frames. From the remaining 13 video sequences we sampled the first 90\% of the frames and use them for training and the remaining 10\% were used for validation. The training set consists of 736 manually-labeled frames and a total of 35,784 frames that we automatically annotate using SegProp.
We divided the dataset in such a way to be representative enough for the variability of different flying scenarios. 

%All our models were trained with RGB frames at a spatial resolution of $2048\times1080$, rescaled from the original 4K resolution of the videos.

\subsection{Comparison with other methods for label propagation}

We did an ablation study in which we measured the performance of our propagation algorithm when we change the propagation length, from 25 frames, up to 100 frames. We performed the study on one of ours clips that was annotated every 25 frames, extending the interpolated results two fold at each step and progressively hiding manually labeled frames, used as ground truth for evaluation. We also compared our results against the SDCNet algorithm proposed in ~\cite{zhu2019improving} that produces state-of-the art results on Cityscapes. We measure mean F-measure over all classes from the selected video and report results in Table ~\ref{table:label_propagation}. While our method alone provides a significant boost over SDCNet, combining the two results in even better results. The combination was done as follows: SDCNet propagation was used, alongside the flow-based and homography-based correspondences within the voting mechanism. Thus SDCNet brought two extra class votes per pixel, one from the left labeled frame and the other from the right one. 
This confirms the intuition that our iterative label propagation procedure 
could take advantage of any accurate procedure that could help in casting votes from the labeled frames to the intermediate unlabeled ones.

Our algorithm performs better than SDCNet in all scenarios, even when a significant number of ground truth frames are missing. When the votes are propagated through 100 frames (2 seconds in our case), the label propagation performance decreases significantly (0.734) but our approach is still better than SDCNet, with the combination giving the best result.

\setlength{\tabcolsep}{4pt}
\begin{table}
\begin{center}
\caption{\label{table:label_propagation} Automatic label propagation comparisons. We measure mean F-measure over all classes. The bolded values are the best results.}
\vspace*{-1mm}
\begin{tabular}{|c||c|c|c|}
\hline
Propagated frames &  SDCNet~\cite{zhu2019improving} & SegProp & SegProp w/ SDCNet~\cite{zhu2019improving}\\
\hline
25  & .834 & .857 & \textbf{.864}\\
50  & .756 & .811 & \textbf{.813}\\
100 & .675 & .728 & \textbf{.734} \\
\hline
\end{tabular}
\end{center}
\vspace{-9mm}%Put here to reduce too much white space after table 
\end{table}
\setlength{\tabcolsep}{1.4pt}

%Comparing our first two interpolation methods numerically, the results don’t exhibit a significant difference. However, looking at the qualitative results, the edges look sharper compared to the sequential variant – in line with the planar patch assumption we made when designing the method. We argue that accurate edges are important for both fast navigation and derived tasks, such as localization.

\subsection{Training scenarios}

Models were trained using the same learning setup. We used Keras deep learning framework with Tensorflow backend. We use RMSprop optimizer with a learning rate starting from 1e-4 and decreasing it, no more than five times when optimization reaches a plateau. Training is done using the early stopping paradigm. We monitor the error on the validation set and suspend the training when the loss has not decayed for 10 epochs. 

In order to assess the gain brought by SegProp we train SafeUAVNet-Large 
in the same training scenario but only on the manually-labeled frames as baseline (termed w/ Base Train in Table \ref{table:experimental_results}). This model was trained using only 736 frames, whilst SafeUAVNet-Large trained w/ SegProp had $\approx 49\times$ more (automatically) annotated frames in addition to the manually labeled ones.
Quantitative results are reported in Table \ref{table:experimental_results}. The overall score was computed as mean F-measure over the whole classes. Some of the classes were not predicted at all by the method w/ Base Train and were marked with .000. The results also show that small classes experience a significant boost, whilst the improvement in the larger ones is smaller. The ambiguity for the land, forest and hill classes is reflected in the results. While sky has the largest score (0.98 F-measure), the residential zones take the second place (0.82 F-measure). We believe improving the latter with temporal coherence constraint or multiple input frames could turn the result into a commercial application.

Qualitative results on our testing set are shown in Figure \ref{fig:qualitative_results}. 
They exhibit good spatial coherency, even though the neural network processes each frame individually. The quality of segmentation is affected by sudden scene geometry changes, cases not well represented in the training videos and and motion blur.

Quantitative results on Ruralscapes, our large dataset with complex and difficult videos,
show that our automatic label propagation algorithm significantly improves segmentation. As expected, for well represented classes we achieve high accuracy, whilst small classes are much harder to segment. Only w/ Base Train option, classes such as person, haystack and car are difficult to detect and missed completely.

%In an bid to increase accuracy, we split the full resolution 4K images into four 2048x1080 px images and performed an additional training.
\vspace{-2mm}
\section{CONCLUSIONS}

We introduced Ruralscapes, the largest high resolution (4K) dataset for dense semantic segmentation in aerial videos from real UAV flights. It will be made publicly available alongside a fast segmentation tool, in a bid to help aerial segmentation algorithms. We proposed an effective iterative label propagation method, SegProp, that requires only a small fraction of labeled frames (about 2 percent in our tests). Our method significantly outperforms SDCNet, the current state-of-the-art in label propagation, in our experiments. We also show that by adding region-wise homographic constraints resulted in sharper edges and overall better segmentations. When combining SegProp with SDCNet
the results improved even further, showing that our voting-based, iterative approach, is general and could work in combination with other propagation methods. Our encouraging experiments demonstrate that
deep neural networks could extensively benefit from the added training labels using the 
proposed label propagation algorithm. Further gains can be achieved by exploring the spatial and temporal coherence from video sequences in order to improve the segmentation result and reduce processing costs, which is especially desirable for on-board UAV processing.

\vspace{5mm}

\noindent\textbf{Acknowledgements} This work was supported by UEFISCDI, under Projects  EEA-RO-2018-0496 and PN-III-P1-1.2-PCCDI-2017-0734. We would also like to express our gratitude to Aurelian Marcu and The Center for Advanced Laser Technologies for providing us GPU training resources.

The code and dataset are available on our website: https://sites.google.com/site/aerialimageunderstanding/.

\noindent\textbf{} 

\bibliographystyle{IEEEtran}
\bibliography{IEEEexample}

\end{document}